%% file: paper.tex
\documentclass[9pt,conference]{IEEEtran}
\ifCLASSINFOpdf
  \usepackage[pdftex]{graphicx}
  \graphicspath{{imgs/}}
\else
\fi
\usepackage{amsmath}
\ifCLASSOPTIONcompsoc
  \usepackage[caption=false,font=normalsize,labelfont=sf,textfont=sf]{subfig}
\else
  \usepackage[caption=false,font=footnotesize]{subfig}
\fi
\usepackage{url}

\usepackage{hyperref}
\usepackage{amssymb}
\usepackage[keeplastbox]{flushend}

\IEEEoverridecommandlockouts

\def\footnoterule{\relax%
  \kern-5pt
  \hbox to \columnwidth{\vrule width 0.4\columnwidth height 0.4pt\hfill}
  \kern4.6pt}

\newif\iffinal
\finaltrue

\begin{document}
%
\title{Cross-Domain Face Verification:\\Matching ID Document and Self-Portrait Photographs}

\iffinal
	\author{\IEEEauthorblockN{Guilherme Folego\,$^{1,2}$\,$^{*}$\thanks{$^{*}$Corresponding author: \tt\href{mailto:gfolego@cpqd.com.br}{gfolego@cpqd.com.br}} \quad
                              Marcus A. Angeloni\,$^{1,2}$ \quad
                              Jos\'{e} Augusto Stuchi\,$^{2,3}$ \quad
                              Alan Godoy\,$^{1,2}$ \quad
                              Anderson Rocha\,$^{2}$} \\
            \IEEEauthorblockA{$^{1}$\,CPqD Foundation, Brazil \qquad $^{2}$\,University of Campinas (Unicamp), Brazil \qquad $^{3}$\,Phelcom Technologies, Brazil}}       
%
\fi


\maketitle

\begin{abstract}
Cross-domain biometrics has been emerging as a new necessity, which poses several additional challenges, including harsh illumination changes, noise, pose variation, among others. In this paper, we explore approaches to cross-domain face verification, comparing self-portrait photographs (``selfies'') to ID documents.
We approach the problem with proper image photometric adjustment and data standardization techniques, along with deep learning methods to extract the most prominent features from the data, reducing the effects of domain shift in this problem. We validate the methods using a novel dataset comprising 50 individuals. The obtained results are promising and indicate that the adopted path is worth further investigation.
\end{abstract}


%
\IEEEpeerreviewmaketitle

%
%
%
%
%

\input{src/1_intro}
\input{src/2_method}
\input{src/3_exp}
\input{src/4_res}
\input{src/5_conc}

\iffinal
\section*{Acknowledgments}
The authors thank Andr\'{e} R. Gon\c{c}alves
for insightful discussions, and for reviewing early versions of this paper. The authors also appreciate the collaboration of all users who helped create the dataset, and the financial support of Capes DeepEyes project and FAPESP D\'{e}j\`{a}Vu Grant \#2015/19222-9.
\fi



\bibliographystyle{IEEEtran}
\bibliography{src/paper}
%
%
%

\end{document}

%% file: src/1_intro.tex

\section{Introduction and Related Work}
\label{sec:intro}

\begin{figure}[!t]
	\centering
	\includegraphics[width=0.85\linewidth]{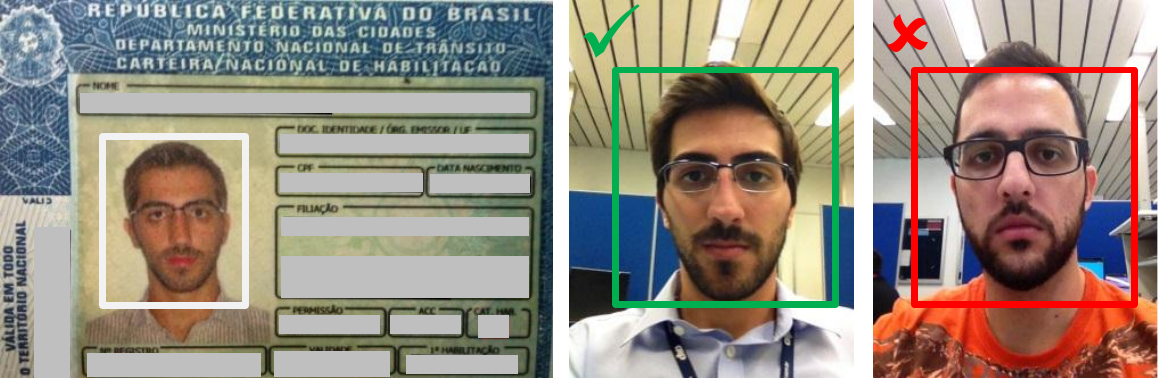}
	\caption{ID document and ``selfie'' verification example.}
	\label{fig:Test2.pdf}
\end{figure}

Currently, a new demand is emerging for facial recognition applications, mainly due to digital life: companies, such as financial institutions, are allowing customers to create accounts using the Internet, without the necessity to go to a physical branch. In this context, the customer ID is required and can be used to check authenticity. This process is usually performed using a photograph of the customer in conjunction with the ID image, as Fig.~\ref{fig:Test2.pdf} illustrates. In this case, a user can capture a ``selfie'' and an ID photograph using a smartphone and send them to the company website, which evaluates the information provided and, if no problem or fraud is detected, creates the user account or requests further action. This is an example of facial recognition systems~\cite{zhao2003face}, which have become increasingly popular in commercial applications in the last decade.
Typically, they are used in security systems and can be combined with other biometric systems, such as fingerprint, iris and voice. 

A generic face recognition system comprises three basic steps: face detection, feature extraction, and recognition.
Face detection methods usually employ Haar-like features, as the ones proposed by Viola \& Jones~\cite{viola2001rapid}, which are efficient and have high detection rates.

Many different techniques are commonly applied in the literature for the feature extraction step. For this purpose, there are mainly two different approaches: handcrafted feature extraction and automatic feature learning.
Handcrafted descriptors are created by human specialists, requiring great effort and knowledge to develop appropriate features to describe the image characteristics.
Examples of these features applied in the face recognition context are
Local Binary Patterns (LBP)~\cite{lbp_ahonen2006face}
and Discrete Cosine Transform (DCT)~\cite{dct_hafed2001face}.
In the automatic feature learning approach, the more distinctive features are determined directly from data~\cite{chen2013blessing}.
Examples of features learned from data in the face recognition scenario are the ones extracted using Deep Belief Networks (DBN)~\cite{dbn_huang2012learning} and Convolutional Neural Networks (CNN)~\cite{facenet}.

The last part of a face recognition system entails the actual recognition step, in which machine learning techniques can be applied in order to classify the extracted features, aiming at recognizing the users. In this case, there are two main use scenarios: identification and verification. 
In our study, we tackle the verification scenario, in which the task is to confirm whether a face image belongs to a specific user whose identity was previously claimed. For this purpose, different classification techniques can be applied, such as Support Vector Machines (SVM)~\cite{svm} and Logistic Regression (LR)~\cite{logisticregression}.

In our work, the biometric system is presented with images obtained from two very different domains: (1) user ``selfies'' obtained under real-world conditions; and (2) a 
photographed image of a driver's license ID. To cope with the main challenges involved in this cross-domain face verification problem, we explored different techniques to enhance the images, diminishing the impacts of illumination changes and domain-shift, and to extract features from the pre-processed data using a pre-trained Convolutional Neural Network. These features were then normalized, combined and used as input for different classifiers. Multiple pipeline combinations were tested aiming at verifying the influence on the final result and the main factors that drive the process.

Our approach differs from the usual ones, as it does not use a gallery or create a biometric profile for each individual from the ``selfie'' or the ID.
By training a classifier to compare images from two specific and different domains, independently of the user, we can specialize on the relation between domains and avoid any need of enrollment process, reducing data needs and privacy issues.

Face recognition and domain adaptation are problems of broad interest for scientific community and there are many publications dealing with them. In~\cite{deepface}, a 3D face modelling method was used to apply a piecewise affine transformation, deriving a face representation using a deep neural network with nine layers. The method was trained using a large dataset with four million labeled images, and the model generalized well for faces in unconstrained environments, even using simple classifiers. The results achieved in the Labeled Faces in The Wild~(LFW) dataset were close to human-level performance.

In~\cite{facenet}, the authors implemented a system that directly learns a mapping from face images onto a compact Euclidean space.
They trained a deep CNN using a very large dataset with $200$ million labeled images. It was used to achieve state-of-the-art results on LFW and YouTube Faces (YTF) datasets, using a small representation of $128$ bytes per face.

In~\cite{vggface}, authors also applied a CNN to extract features, but this network was trained using a smaller dataset, with about $2.6$ million labeled images. It achieved results comparable to state-of-the-art on LFW and YTF. In particular, one of the trained networks was made publicly available, favoring tests, comparisons and improvements.

Classification algorithms often have a performance drop under cross-domain conditions (\textit{e.g.}, digitized face images vs. live pictures). This is easily explained as images may vary in blur, illumination, alignment, noise, or facial expression. In this regard, the method proposed in~\cite{domainadaptation1} deals with this cross-domain problem by deriving a latent subspace, characterizing the multifactor variations. Images were synthesized in order to produce different illumination and blur conditions, and other 2D perturbations, forming a tensor to represent the face. Results indicated that the method is effective on constrained and unconstrained datasets. However, different from our work, this method did not deal with heterogeneous modalities of pictures, \textit{i.e.}, digitized printed documents \textit{vs.} live pictures.

To the best of our knowledge there is no academic work dealing specifically with the user ``selfie'' and ID image verification, although there are some commercial solutions, such as Netverify~\cite{jumio}.




%% file: src/2_method.tex

\section{Proposed Method}
\label{sec:method}

\begin{figure}[!t]
	\centering
	\includegraphics[width=0.70\linewidth]{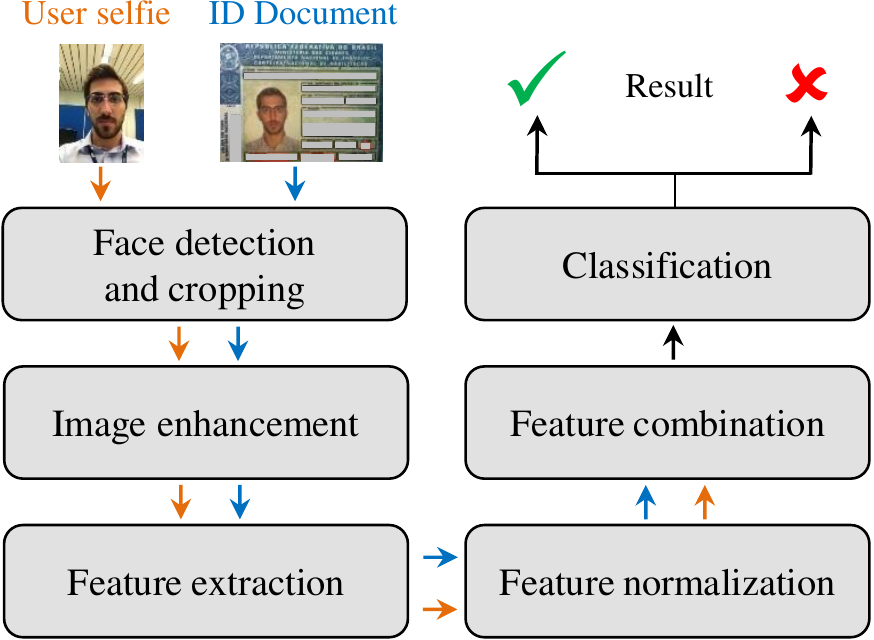}
	\caption{Pipeline overview for ``selfie'' and ID images verification.}
	\label{fig:Flow.pdf}
\end{figure}

Here, we detail the pipeline (see Fig.~\ref{fig:Flow.pdf}) used to evaluate different approaches for each step of a cross-domain face recognition system.


\subsection{Face detection and cropping}
As face detection is not our focus, we adopt a traditional method~\cite{viola2001rapid} for both input images in all evaluations. For a more uniform detection, and to include the outer part of the ear, chin or hair, each side of the detected region of interest is expanded by $22\%$.
The eye-coordinates are used to perform a geometric normalization through a planar rotation to position the eyes in a zero degree angle. Finally, the face images are downsized to $224\!\times\!224$ pixels using bilinear interpolation. Fig.~\ref{fig:Test2.pdf} shows some examples of this step. 

\subsection{Image enhancement}
\begin{figure}[!t]
	\centering
	\subfloat[Original]{\includegraphics[width=0.15\linewidth]{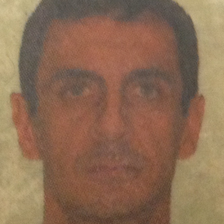} %
	\label{sample1}}
	\qquad
	\subfloat[Retinex]{\includegraphics[width=0.15\linewidth]{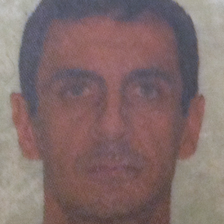}}
	\qquad
	\subfloat[ACE]{\includegraphics[width=0.15\linewidth]{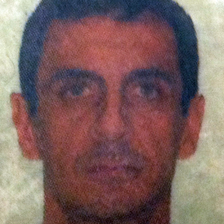}}
	\qquad
	\subfloat[CLAHE]{\includegraphics[width=0.15\linewidth]{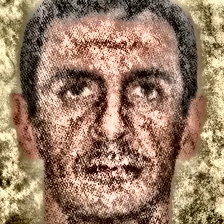} %
	\label{sample2}}
	\qquad

	\subfloat[Original]{\includegraphics[width=0.15\linewidth]{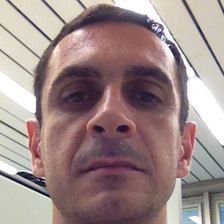} %
	\label{sample3}}
	\qquad
	\subfloat[Retinex]{\includegraphics[width=0.15\linewidth]{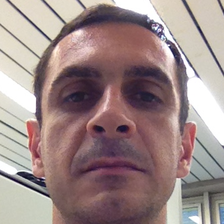}}
	\qquad
	\subfloat[ACE]{\includegraphics[width=0.15\linewidth]{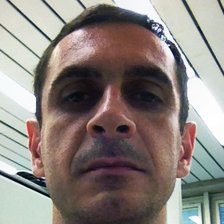}}
	\qquad
	\subfloat[CLAHE]{\includegraphics[width=0.15\linewidth]{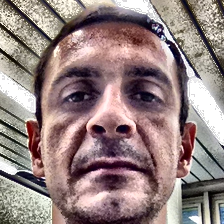} %
	\label{sample4}}
	\qquad

	\caption{Photometric enhancement methods applied to an ID document picture \textit{\protect\subref{sample1}-\protect\subref{sample2}},
	and to a user ``selfie'' picture \textit{\protect\subref{sample3}-\protect\subref{sample4}}.}
	\label{fig:samples_face}
\end{figure}

One of the most prominent problems when comparing digitized documents to live pictures
is the serious illumination effects due to the domain shift, along with noise amplification. To mitigate such effects, we evaluate three algorithms (see Fig.~\ref{fig:samples_face} for their results).  

The first one is based on the Retinex theory of visual color constancy~\cite{hist:retinex}, which argues that perceived white is associated with the maximum cone signals of the human visual system.
It reduces the variations in color intensities from the different domains.
The second is the Automatic Color Equalization (ACE)~\cite{hist:ace}, which is based on a computational model of the human visual system that equalizes simultaneously global and local effects. It obtains good contrast enhancements, which is an effort to approximate the two distinct sources of input images.
The last assessed method is the Contrast Limited Adaptive Histogram Equalization (CLAHE)~\cite{hist:clahe}, which divides the input image into small blocks, applies a conventional histogram equalization in each block, and then checks if any histogram bin is above the contrast limit. This way, the method avoids over-brightness situations and noise amplification~\cite{hist:clahe},
typical when capturing printed images. For all these techniques, the default configurations were used.

\subsection{Feature extraction}


To extract complex features directly from the images we employ a CNN-based transfer learning approach, a technique that has achieved good results in many different computer vision applications~\cite{cnnofftheshelf}. We computed descriptors using the VGG-Face network~\cite{vggface}. In this work, a number of networks were trained using a dataset with more than $2.6$ million face images of $2{,}622$ different identities and were able to achieve state-of-the-art results in LFW and YTF datasets. In particular, the weights of the trained network configuration \textit{D}~\cite{vgg} have been made publicly available. This network  takes $224\!\times\!224$-pixel RGB images as input. It has $16$ weight layers and $138$ million parameters, which makes it an interesting option for extracting complex features from images of faces.

By using transfer learning with a pre-trained CNN, we can benefit from visual properties especially relevant to face recognition that are more robust than handcrafted features, avoiding the large dataset (in the order of millions of images) with ``selfies'' and IDs that would be needed to train a deep network specific to the problem in hand.





Works that employ CNN-based transfer learning techniques usually extract features from the second-to-last layer, which, in this network, is internally called \textit{fc7}~\cite{vggface,vgg}, and has $4{,}096$ dimensions. In our work, we explore two additional layers that are close to \textit{fc7}, namely, the third-to-last layer (\textit{fc6}) with $4{,}096$ dimensions, and the last layer (\textit{fc8}) before softmax activation with $2{,}622$ dimensions.


Considering that the activation functions after \textit{fc6} and \textit{fc7} are Rectified Linear Units (ReLU), which are defined as $f(x) = \max(0, x)$, these layers tend to produce sparse outputs. Even though this may lead to compact and meaningful representations, it might eventually discard important information.
As an alternative, 
we also independently analyze the features from \textit{fc6} and \textit{fc7} before the activation function. 
However, since the network has never seen negative values while training, it is not possible to hypothesize whether such non-sparse features will be actually useful or rather just behave like random values.
It is important to note that features are extracted individually from ID and ``selfie'' images, and that removing the activation does not alter the output dimension of the layer.

\subsection{Feature normalization}
Given the cross-domain (heterogeneous sources) setup, features extracted in each domain might have significantly different magnitude ranges. Therefore, we analyze three different approaches for feature normalization.

Consider the $p$-norm of a feature vector $\mathbf{x} \in {\mathbb{R}}^{n}$, given by
$
{\lVert\mathbf{x}\rVert}_p = {\left( \sum_{i=1}^{n} {\lvert {x}_i \rvert}^p  \right)}^{1/p}
$.
The L1-normalized feature vector is given by
$
\hat{\mathbf{x}} = \mathbf{x} / {\lVert\mathbf{x}\rVert}_1
$ and the L2, by
$
\hat{\mathbf{x}} = \mathbf{x} / {\lVert\mathbf{x}\rVert}_2
$. However, since these normalization techniques simply divide the original feature vector by a scalar, they maintain the original sparsity. One option to reduce sparsity is to apply the Z-score normalization.
Given a feature vector $\mathbf{x}$, its mean $\mu$, and its standard deviation $\sigma$, the Z-normalized feature vector is given by
$
\hat{\mathbf{x}} = (\mathbf{x} - \mu) / \sigma
$.

The main difference between L1 and L2-normalization is that larger values are more emphasized in L2 than in L1. On the other hand, Z-normalization standardizes vectors to zero mean and unit variance, which might be appropriate for domain-shift conditions. 

\subsection{Feature combination}
Let $\mathbf{a}$ and $\mathbf{b}$ be the $d$-dimensional feature vectors of an ID and a ``selfie'', respectively.
Before proceeding with the classification, we need to combine them in some way in order to emphasize their different properties. The trivial option is to simply concatenate these features, however, our preliminary experiments indicate that this approach does not produce meaningful features, and it also doubles the size of the final feature vector. Thus, we evaluate four methods to calculate the similarity of two vectors and, at the same time, keep the original dimension $d$. The four analyzed techniques, resulting in the final feature vector $\mathbf{f}$, are:

\begin{itemize}
\item Absolute value of subtraction:
$
\mathbf{f} = \lvert \mathbf{a} - \mathbf{b} \rvert
$

\item Element-wise multiplication:
$
\mathbf{f} = \mathbf{a} \circ \mathbf{b} \implies {f}_i = {a}_i \cdot {b}_i ,$
$i = 1,\dotsc,d
$

\item Cross-correlation~\cite{pratt1991digital}:
$
{f}_i = \sum_{j=-\infty}^{\infty} {a}_i \cdot {b}_{i+j} ,\ i\!=\!1,\dotsc,d
$,
 where indices that are out of range are set to $0$.

\item Phase correlation~\cite{pratt1991digital}:
$
\mathbf{f} = \operatorname{IDFT}\left({ \mathbf{G} / {\lVert\mathbf{G}\rVert}_2   }\right)
$, where
$
\mathbf{G} = \operatorname{DFT}(\mathbf{a}) \circ \operatorname{DFT}(\mathbf{b})
$,
$\operatorname{IDFT}$ is the Inverse Discrete Fourier Transform, and $\operatorname{DFT}$ is the Discrete Fourier Transform.

\end{itemize}

When vectors $\mathbf{a}$ and $\mathbf{b}$ are similar, the absolute value of subtraction should produce smaller features, while element-wise multiplication, cross-correlation and phase correlation should produce larger features.
The element-wise multiplication is similar to the cosine distance, but it enables the classifier to learn a more specific decision function based on the feature values, instead of simply summing them up.

\subsection{Classification}
In the classification step, we evaluate three techniques: Support Vector Machine (SVM)~\cite{svm} with linear decision function and with radial basis function (RBF) kernels, and Logistic Regression (LR)~\cite{logisticregression}. 

SVM is a discriminative classifier that aims to construct a maximum-margin separating hyperplane. In both Linear and RBF SVM, we tune the hyperparameters $C$, which controls the cost of misclassifications, and the class weights $W$, which controls the balance between class frequencies. For the RBF SVM, we also optimize the dispersion  $\gamma$, which is related to the nonlinear similarity measure on the high dimensional feature mapping. 

LR is a probabilistic classifier that fits a sigmoid function to estimate the likelihood of a given data point belonging to a specific class. Its penalty norm and regularization ($C$) parameters are optimized during training.

All hyperparameters are optimized with a grid search, and their respective range values are
$C={2}^{-25},\dotsc,{2}^{10}$, $\gamma={2}^{-25},\dotsc,{2}^{10}$, penalty norm $l1$ or $l2$, and $W$ set to either equal class weights, or weights inversely proportional to class frequencies.

\subsection{Pipeline optimization}

In order to find the combination that gives the best result, we perform a greedy pipeline optimization, in which we select the technique with dominating performance in each step.
We understand that this approach may not determine the best overall result, which would require a grid search on all possible combinations, but we believe to be a good compromise between efficiency and effectiveness.

Additionally, we carry out some statistical tests to calculate the significance of our findings.
In each step, we perform the Kruskal-Wallis test for stochastic dominance among groups~\cite{kruskal} to find out whether there is at least one technique with significant difference from the others.
Then, if the test is positive, we proceed with multiple pairwise comparisons using Dunn's \textit{post hoc} test for stochastic dominance~\cite{dunn} to identify which pairs of techniques are statistically different. It is important to note that we minimize false discoveries by controlling the Family-Wise Error Rate with Bonferroni adjustment, which multiplies the calculated p-values by the number of tests performed.
We consider a result statistically significant if its test's p-value is $\leq 0.05$ (\textit{i.e.}, $95\%$ confidence).

%% file: src/3_exp.tex

\section{Experimental Setup}
\label{sec:exp}


\subsection{Dataset}

Since no public database is available for the cross-domain problem of face verification using ID documents and ``selfie'' photographs, we created a novel one, named
\iffinal
CPqD Driver's License Database.
\fi
To this end, we have chosen the Brazilian Driver's License as ID, since it is a national document that follows a unique template, must be updated periodically (every five years), and people often carry it with them.

\begin{figure}[!t]
	\centering
	\includegraphics[width=0.75\linewidth]{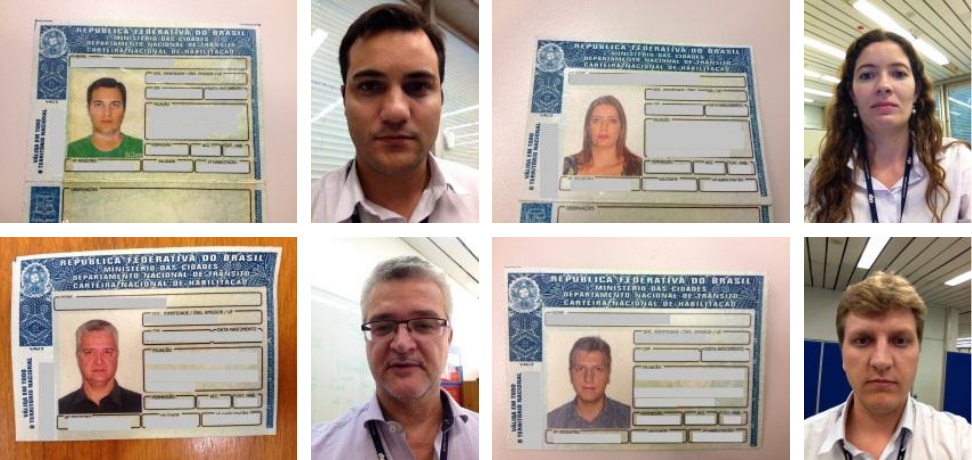}
	\caption{Samples from
	\iffinal
	CPqD Driver's License Database.
	\fi
	}
	\label{fig:database.pdf}
\end{figure}

This dataset comprises pictures taken by $50$ users with an iPhone~4S, using the standard camera application of the iOS (see Fig.~\ref{fig:database.pdf} for some examples). There are $13$ females and $37$ males, collected in an office environment. Ideally, we would have several different locations and hardwares, so the recognition system would learn to ignore such differences. However, having a single standard is an option to isolate potential factors that might influence our results. Additionally, we understand that our dataset is small, but we emphasize that it is quite difficult to collect such private and sensitive data from users.


Given the defined setting, each user was responsible for taking their self-portrait (``selfies'') and ID photographs, thus inserting an inherent variation due to ergonomics. Users were only instructed to capture a ``selfie'' in portrait orientation using the front camera ($480\!\times\!640$ pixels), and their ID document (without the protective plastic sleeve) in landscape orientation with the main camera ($3{,}264\!\times\!2{,}448$ pixels).

\subsection{Metrics}

In this paper, we adopt traditional measures used in biometrics:
False Acceptance Rate ($\texttt{FAR}$), False Rejection Rate ($\texttt{FRR}$), Equal Error Rate ($\texttt{EER}$), and Half Total Error Rate ($\texttt{HTER}$).

The False Acceptance Rate ($\texttt{FAR}$) is defined as the fraction of impostor trials incorrectly accepted by the system, \textit{i.e.}, $\texttt{FAR}=\texttt{FP}/(\texttt{FP}\!+\!\texttt{TN})$. Conversely, the False Rejection Rate ($\texttt{FRR}$) is the fraction of genuine individuals wrongly rejected by the system, \textit{i.e.}, $\texttt{FRR}=\texttt{FN}/(\texttt{FN}\!+\!\texttt{TP})$. 

A broadly used measure to summarize the performance of a system is the Equal Error Rate ($\texttt{EER}$), defined as the error rate at the threshold where $\texttt{FAR} = \texttt{FRR}$.
Finally, another possible way to measure the performance is the Half Total Error Rate ($\texttt{HTER}$), which also combines $\texttt{FAR}$ and $\texttt{FRR}$ into a single measure, enabling the comparison of systems. This measure is calculated at a given threshold $\tau$ (calculated during training) and is defined as $\texttt{HTER}(\tau) = (\texttt{FAR}(\tau) + \texttt{FRR}(\tau))/2$. 

$\texttt{HTER}$ is often used to compare models in person authentication and it represents a particular case of a decision cost function where the costs of a false acceptance and a false rejection are the same~\cite{bengio2004statistical}.

\subsection{Evaluation protocol}
The evaluation protocol starts with a random split of the users into three disjoint sets: training, development, and evaluation.
The training set contains $60\%$ ($30$ users), while each development and evaluation sets have $20\%$ ($10$ users each).
In each set, we generate all possible pairs of IDs and ``selfies''. Thus, there are $30$ positive samples and $870$ negative samples in the training set, whereas there are $10$ positive and $90$ negative samples in each development and evaluation sets.

To tune hyperparameters for the classification model, we use a cross-validation scheme splitting the users in the training set into three disjoint folds. Therefore, in each fold, there are $10$ users, and all possible pairs are generated, thus resulting in $10$ positive samples and $90$ negative samples per fold. After the best set of hyperparameters is found using the average $\texttt{EER}$ as comparison metric, we use the complete training set with all $900$ samples to train the final classifier.

We thus calculate the scores for each pair in the development set, which are then used to calculate the $\texttt{EER}$ threshold. Finally, this threshold is used to calculate the $\texttt{HTER}$ in the evaluation set.

To reduce dependency on initial dataset split, we repeat this entire procedure with $100$ different random splits for the training, development and evaluation sets.
For a fair comparison among different approaches, the same $100$ splits are used in all experiments.

\subsection{Baselines}

To better assess that our results are not simply due to chance, we consider three straightforward baseline methods. If a classifier always predicts as positive (match), the corresponding $\texttt{FAR}$ is $1.0$, and $\texttt{FRR}$ is $0.0$, resulting in $\texttt{HTER}\!=\!0.5$. Conversely, if a classifier always predicts as negative (mismatch), the corresponding $\texttt{FAR}$ is $0.0$, and $\texttt{FRR}$ is $1.0$, resulting in $\texttt{HTER}\!=\!0.5$. Finally, a random classifier would yield an $\texttt{FAR} = \texttt{FRR} = \texttt{HTER} = 0.5$.

To determine the best technique in each step, we used the following initial configuration:
no image enhancement, features extracted from the CNN's \textit{fc7} with ReLU layer, no feature normalization, combination with absolute value of subtraction, and a Linear SVM classifier.

%% file: src/4_res.tex

\section{Results and discussions}
\label{sec:res}


We now turn our attention to the experimental results. 
In a given step, we choose the technique providing the best median $\texttt{HTER}$ value, and then proceed with the optimization of the following step.
We also perform a nonparametric statistical test to measure the confidence of such choices.

\begin{figure}[!t]
	\centering
	\includegraphics[width=1\linewidth]{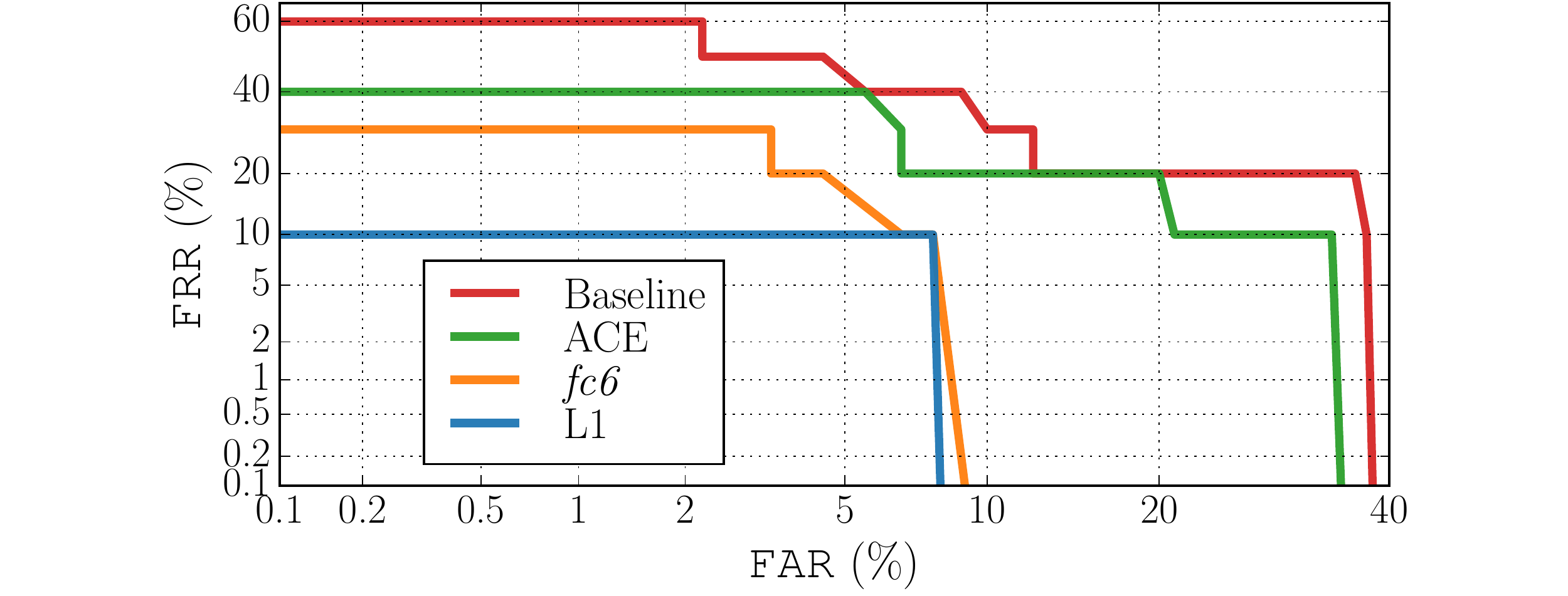}
	\caption{Detection-Error Tradeoff curves for the baseline pipeline and each step that improved the results.}
	\label{fig:det.pdf}
\end{figure}

To better illustrate the improvement in each step of the pipeline, in Fig.~\ref{fig:det.pdf} we show Detection-Error Tradeoff (DET) curves~\cite{martin1997det} for each step that improved the median $\texttt{HTER}$.
Since we have $100$ executions per configuration, we plot results for a execution selected randomly among those in the median $\texttt{HTER}$ value.

\subsection{Image enhancement}
We start answering the following question: Does image enhancement reduce the impact of domain-shift for this problem and improve the results?
Table~\ref{tab:hist} shows the results for each method (where \textit{None} is the baseline). The corresponding Kruskal-Wallis p-value is $0.00028$, clearly indicating that there is a statistical difference amongst these methods. To better determine such difference, we also report the Dunn's test results.

\begin{table}[h!]
\centering

\caption{Results for image enhancement: $\texttt{HTER}$ statistics (top), \qquad \qquad and Dunn's test p-values (bottom).}
\label{tab:hist}
\begin{tabular}{c*{4}{|c}}
\textbf{Method} & \textbf{Median} & \textbf{Mean} $\pm$ \textbf{StdDev} & \textbf{Min} & \textbf{Max} \\ \hline
None	& $0.22500$ & $0.23667 \pm 0.07943$ & $0.02222$ & $0.47778$ \\
Retinex	& $0.22778$ & $0.23722 \pm 0.07657$ & $0.06111$ & $0.48333$ \\
ACE		& $0.20833$ & $0.21822 \pm 0.06757$ & $0.06667$ & $0.40000$ \\
CLAHE	& $0.25278$ & $0.26711 \pm 0.07693$ & $0.10556$ & $0.46667$ \\
\end{tabular}

\bigskip

\begin{tabular}{c|*{3}{c}}
\textbf{Method}	& None	 	& Retinex	&	ACE \\ \hline
Retinex	& $1.00000$ & --- 		& --- \\
ACE		& $0.23989$ & $0.23973$ & --- \\
CLAHE	& $0.03091$ & $0.03094$ & $0.00005$
\end{tabular}

\end{table}

From this table, we are able to infer that CLAHE had the worst performance, which was statistically significant. Also, Retinex performed similarly to the original pictures, while ACE was slightly better than these two, but such differences were not statistically significant.
Thus, we set ACE as our image enhancement method for diminishing the impacts of domain-shift changes in illumination, and present the respective DET curve in Fig.~\ref{fig:det.pdf}.

As Fig.~\ref{fig:samples_face} depicts, both Retinex and ACE correct for the color deviation in the original image, however, it is noticeable how ACE increases contrast. Even though CLAHE should also adjust for contrast, the default configuration exceeds in this adjustment, making the method perform worse.

\subsection{Feature layers}
We now focus on the question: Which network layer provides a better generalization for this problem?
Table~\ref{tab:feat} shows these results, where we append the letter \textit{n} to indicate that the ReLU layer has been removed.
The Kruskal-Wallis p-value found was $0.00000$, showing statistical difference among the methods.

\begin{table}[h!]
\caption{Results for feature layers: $\texttt{HTER}$ statistics (top), \qquad \qquad \qquad and Dunn's test p-values (bottom).}
\label{tab:feat}
\centering
\begin{tabular}{c*{4}{|c}}
\textbf{Layer} & \textbf{Median} & \textbf{Mean} $\pm$ \textbf{StdDev} & \textbf{Min} & \textbf{Max} \\ \hline
\textit{fc6n}	& $0.10278$ & $0.11100 \pm 0.05961$ & $0.00000$ & $0.28889$ \\
\textit{fc6}	& $0.10000$ & $0.10900 \pm 0.05937$ & $0.01111$ & $0.35000$ \\
\textit{fc7n}	& $0.12222$ & $0.13078 \pm 0.06424$ & $0.00556$ & $0.31111$ \\
\textit{fc7}	& $0.20833$ & $0.21822 \pm 0.06757$ & $0.06667$ & $0.40000$ \\
\textit{fc8}	& $0.16111$ & $0.18339 \pm 0.07638$ & $0.03889$ & $0.37778$ \\
\end{tabular}

\bigskip

\begin{tabular}{c|*{4}{c}}
\textbf{Layer}	& \textit{fc6n}	 	& \textit{fc6}	&	\textit{fc7n}		& \textit{fc7} \\ \hline
\textit{fc6}	& $1.00000$ & --- 		& ---		& --- \\
\textit{fc7n}	& $0.18390$ & $0.07790$ & --- 		& --- \\
\textit{fc7}	& $0.00000$ & $0.00000$ & $0.00000$ & --- \\
\textit{fc8}	& $0.00000$ & $0.00000$ & $0.00003$ & $0.00936$
\end{tabular}
\end{table}

These values evince no clear difference between \textit{fc6} with and without ReLU. However, in our experiments, using either the traditional \textit{fc7} or even \textit{fc8} are statistically worse. This indicates that researches employing CNN-based transfer learning techniques should take other layers into consideration as well, instead of only the traditional one right before the last fully-connected layer (\textit{i.e.}, \textit{fc7}). The reason for this finding is possibly due to the nature of the network. Given that the used network was trained for face recognition on a single domain, possibly these last layers are too domain specialized, lacking the generalization power for a domain-shift problem such as ours.

Moreover, the features extracted from \textit{fc6} are $80.5\%$ sparse, and from \textit{fc7} are $90.6\%$, while \textit{fc6n}, \textit{fc7n} and \textit{fc8} are all dense. It is interesting to note that removing the ReLU activation from \textit{fc6} barely changed the results, whereas the outcomes from \textit{fc7} were significantly better when not using the activation. This indicates that the negative values in \textit{fc7n} were rather meaningful, but the ones from \textit{fc6n} were not, and that the classifier was able to select the important features from \textit{fc6n}, making it perform similarly to \textit{fc6}.

Following our greedy pipeline optimization technique, we select the \textit{fc6} layer herein.
We present the respective DET curve in Fig.~\ref{fig:det.pdf}.

\subsection{Feature normalization}

In this step, we ask the following question: Does normalizing the features help mitigating the feature magnitude differences between the different domains? Table~\ref{tab:norm} shows the results. Additionally, the Kruskal-Wallis p-value was $0.00000$, showing that normalization techniques performed differently.

\begin{table}[h!]
\caption{Results for feature normalization: $\texttt{HTER}$ statistics (top), \qquad and Dunn's test p-values (bottom).}
\label{tab:norm}
\centering
\begin{tabular}{c*{4}{|c}}
\textbf{Method} & \textbf{Median} & \textbf{Mean} $\pm$ \textbf{StdDev} & \textbf{Min} & \textbf{Max} \\ \hline
None	& $0.10000$ & $0.10900 \pm 0.05937$ & $0.01111$ & $0.35000$ \\
L1		& $0.06111$ & $0.07283 \pm 0.04940$ & $0.00000$ & $0.25556$ \\
L2		& $0.06667$ & $0.07022 \pm 0.04115$ & $0.00000$ & $0.23333$ \\
Z		& $0.07222$ & $0.08500 \pm 0.05404$ & $0.00556$ & $0.26667$ \\
\end{tabular}

\bigskip

\begin{tabular}{c|*{3}{c}}
\textbf{Method}	& None	 	& L1	&	L2 \\ \hline
L1	& $0.00000$ & --- 		& --- \\
L2	& $0.00000$ & $1.00000$ & --- \\
Z	& $0.00438$ & $0.27045$ & $0.23409$
\end{tabular}
\end{table}

From these results, we see that any of the analyzed normalization methods significantly outperforms the baseline, which lacks feature normalization. Both L1 and L2 techniques behave similarly, having Z-normalization as a good third option.
Thus, we set L1 normalization for our pipeline, and plot the corresponding DET curve in Fig.~\ref{fig:det.pdf}.

\subsection{Feature combination}

We now tackle the fourth research question: Which form of feature combination (for matching verification and pair learning) of each ID and ``selfie'' image is more appropriate for our problem? Table~\ref{tab:comb} shows the results. In this table, \textit{Sub} stands for absolute value of subtraction (used as baseline), \textit{Mult} stands for element-wise multiplication, \textit{Cross} for cross-correlation, and \textit{Phase} stands for phase correlation. The p-value from the Kruskal-Wallis' test was $0.00000$.

\begin{table}[h!]
\caption{Results for feature combination: $\texttt{HTER}$ statistics (top), \qquad \qquad and Dunn's test p-values (bottom).}
\label{tab:comb}
\centering
\begin{tabular}{c*{4}{|c}}
\textbf{Method} & \textbf{Median} & \textbf{Mean} $\pm$ \textbf{StdDev} & \textbf{Min} & \textbf{Max} \\ \hline
Sub		& $0.06111$ & $0.07283 \pm 0.04940$ & $0.00000$ & $0.25556$ \\
Mult	& $0.16111$ & $0.17128 \pm 0.05990$ & $0.06111$ & $0.34444$ \\
Cross	& $0.08333$ & $0.08528 \pm 0.04674$ & $0.00556$ & $0.26111$ \\
Phase	& $0.07778$ & $0.08289 \pm 0.04984$ & $0.00556$ & $0.26111$ \\
\end{tabular}

\bigskip

\begin{tabular}{c|*{3}{c}}
Method	& Sub	 	& Mult	&	Corr \\ \hline
Mult	& $0.00000$ & --- 		& --- \\
Corr	& $0.17969$ & $0.00000$ & --- \\
Phase	& $0.43391$ & $0.00000$ & $1.00000$
\end{tabular}
\end{table}


Clearly, the element-wise multiplication results in the worst performance.
Considering that the different feature combination techniques did not improve on our baseline, the corresponding DET curve is the same as before, and, as such, a new curve is not plotted.

Considering that the linear SVM is used for classification, our intuition is that both cross-correlation and phase correlation features are not as linearly separable as those from the subtraction, resulting in a lack of improvement.
Additionally, the element-wise multiplication should increase the feature sparsity, loosing important information, thus resulting in a statistically worse performance.

\subsection{Classification}
Finally, we ask the last research question: Does using different classifiers improve the results?
Table~\ref{tab:class} shows the results for this experiment.
Since the respective resulting Kruskal-Wallis p-value was $0.96365$, we can conclude that none of the tested classifiers led to a statistically different performance, showing that the used features are already reasonably linearly separable, thus, not needing a nonlinear discriminant function. Due to the lack of improvement, we do not plot such results in Fig.~\ref{fig:det.pdf}.

\begin{table}[h!]
\caption{Results for classifiers: $\texttt{HTER}$ statistics.}
\label{tab:class}
\centering
\begin{tabular}{c*{4}{|c}}
\textbf{Method} & \textbf{Median} & \textbf{Mean} $\pm$ \textbf{StdDev} & \textbf{Min} & \textbf{Max} \\ \hline
Linear SVM	& $0.06111$ & $0.07283 \pm 0.04940$ & $0.00000$ & $0.25556$ \\
RBF SVM		& $0.06667$ & $0.07356 \pm 0.05143$ & $0.00000$ & $0.25000$ \\
LR			& $0.06667$ & $0.07233 \pm 0.04515$ & $0.00000$ & $0.22222$ \\
\end{tabular}
\end{table}

\subsection{Summary}

\begin{figure}[!t]
	\centering
	\includegraphics[width=1\linewidth]{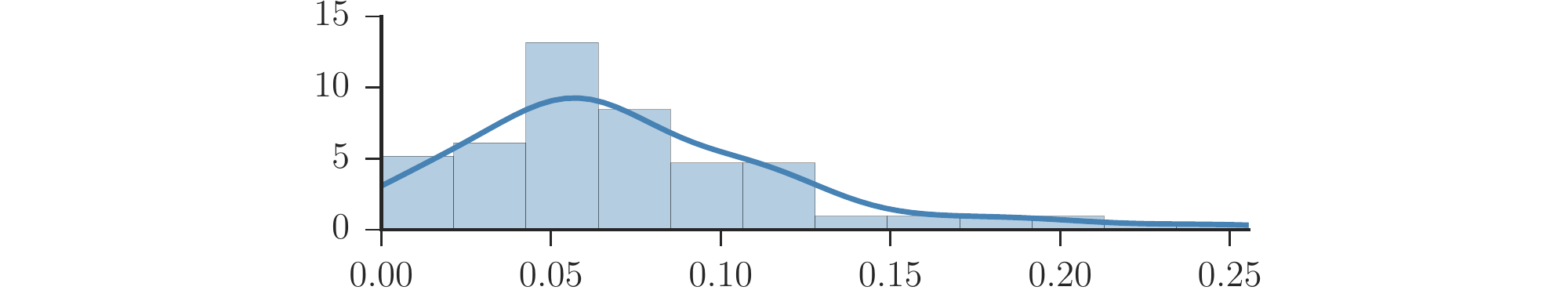}
	\caption{Density histogram of $\texttt{HTER}$ values from our best pipeline. Most of the results are around 0.06, which represents a low error for this cross-domain problem. To put things in perspective, the initial baseline (Table~\ref{tab:hist}) presented a median $\texttt{HTER}$ of $0.22$.}
	\label{fig:hist.pdf}
\end{figure}

Our optimization approach yielded the following pipeline: image enhancement with ACE, features extracted from the CNN's \textit{fc6} with ReLU, L1 normalization, combination with absolute value of subtraction, and Linear SVM classifier. The median $\texttt{HTER}$ is $0.06111$, with corresponding $\texttt{FAR}\!=\!0.02222$ and $\texttt{FRR}\!=\!0.10000$.
In other words, this model correctly authenticates $90\%$ of legitimate attempts, while rejecting impostors with an $\texttt{FAR}$ around $2\%$, a remarkable first result for cross-domain authentication.
Considering this pipeline, we present the density histogram and a kernel density estimation from the $100$ experiments (with random dataset splits) in Fig.~\ref{fig:hist.pdf}.

%% file: src/5_conc.tex

\section{Conclusions}
\label{sec:conc}

In this paper, we investigated the face verification problem on a challenging cross-domain setup, comparing images of a ``selfie'' from a user in the real world and photographs from printed driver's license IDs. This is a new demand that can be employed to many practical scenarios nowadays, and a problem that has not received attention from the academia thus far.
The proposed method analyzed different approaches for image enhancement, feature extraction, normalization and combination, and for classification. In this process, we evaluated $16$ pipeline combinations, generating a total of $1{,}600$ models.

Given the performed experiments, interesting results were found and represent
the main contributions of this paper. First of all, under cross-domain conditions, image enhancement (photometric compensation) methods can improve results.
Second, researches applying CNN-based transfer learning approaches to extract features should explore the network of interest, as different layers and activations can make statistically significant improvements, specially when dealing with cross-domain setups. The earlier layers in the network may provide more general features than those specialized ones of the very last layers of the network. Third, this same reasoning also applies to feature normalization, considering that all analyzed methods had superior performance than the lack of normalization. When assessing features from diverse domains, it is expected that they have different magnitudes and, hence, tackling such difference is a valid concern. 
Finally, different feature combination and classification techniques were evaluated, but, in our specific problem, they did not improve the results. Given all the different combinations that were tested here, we achieved promising results, which represent a considerable starting point for researchers interested in authenticating people under a cross-domain (user ``selfie'' \textit{vs.} ID document) setup.


Future research includes performing a grid search of the complete pipeline, instead of optimizing each step greedy-wise. This could also include the independent optimization for ID and ``selfie'' images, rather than sharing the same image enhancement, feature layer, and feature normalization methods.
Greatly increasing the dataset is mandatory to better evaluate the proposed pipelines and to train more accurate classifiers.
In terms of CNNs, fine tuning the weights in a pre-trained network is worth investigating, and exploring other architectures, such as siamese networks~\cite{siamese}, also holds promise. However, either solution requires training a CNN and, therefore, needs a large amount of data, or smarter optimization approaches~\cite{cnn_few_samples}.
$\blacksquare$
\\